\documentclass[letterpaper]{article} 
\usepackage{aaai25}  
\usepackage{times}  
\usepackage{helvet}  
\usepackage{courier}  
\usepackage[hyphens]{url}  
\usepackage{graphicx} 
\usepackage{natbib}  
\usepackage{caption} 
\usepackage{algorithm}
\usepackage{algorithmic}
\usepackage{amsmath}
\usepackage{amssymb}
\usepackage{threeparttable}
\usepackage{booktabs}
\usepackage{multirow} 
\usepackage[table]{xcolor}
\usepackage{newfloat}
\usepackage{listings}
\DeclareCaptionStyle{ruled}{labelfont=normalfont,labelsep=colon,strut=off} 
\floatstyle{ruled}
\newfloat{listing}{tb}{lst}{}
\floatname{listing}{Listing}
\title{Hyperbolic-constraint Point Cloud Reconstruction from Single RGB-D Images}
\author{
    Wenrui Li \textsuperscript{\rm 1}, Zhe Yang \textsuperscript{\rm 2}, Wei Han \textsuperscript{\rm 1}, Hengyu Man \textsuperscript{\rm 1}, Xingtao Wang \textsuperscript{\rm 1}\textsuperscript{\rm 3}\thanks{Corresponding author.} and Xiaopeng Fan \textsuperscript{\rm 1}\textsuperscript{\rm 3}\textsuperscript{\rm 4}\\
}
\affiliations{
    \textsuperscript{\rm 1}Harbin Institute of Technology \textsuperscript{\rm 2}University of Electronic Science and Technology of China\\ \textsuperscript{\rm 3}Harbin Institute of Technology Suzhou Research Institute \textsuperscript{\rm 4}Peng Cheng Laboratory

    liwr618@163.com; xiazhe200006@gmail.com; 2021111641@stu.hit.edu.cn;\\ manhengyu@hotmail.com; xtwang@hit.edu.cn; fxp@hit.edu.cn 
}

\usepackage{bibentry}

\begin{document}

\maketitle

\begin{abstract}
Reconstructing desired objects and scenes has long been a primary goal in 3D computer vision. Single-view point cloud reconstruction has become a popular technique due to its low cost and accurate results. However, single-view reconstruction methods often rely on expensive CAD models and complex geometric priors. Effectively utilizing prior knowledge about the data remains a challenge. In this paper, we introduce hyperbolic space to 3D point cloud reconstruction, enabling the model to represent and understand complex hierarchical structures in point clouds with low distortion. We build upon previous methods by proposing a hyperbolic Chamfer distance and a regularized triplet loss to enhance the relationship between partial and complete point clouds. Additionally, we design adaptive boundary conditions to improve the model's understanding and reconstruction of 3D structures. Our model outperforms most existing models, and ablation studies demonstrate the significance of our model and its components. Experimental results show that our method significantly improves feature extraction capabilities. Our model achieves outstanding performance in 3D reconstruction tasks.
\end{abstract}

%

\section{Introduction}
With the advancement of deep learning \cite{tang01,tang02,chen01,chen02}, technologies such as 3D computer vision \cite{b5,b6} and multimodality \cite{wenrui1,wenrui2,wenrui3,wenrui4} have garnered widespread attention. Point cloud reconstruction is now a leading technique in 3D computer vision, valued for its precise positional information and detailed reconstruction capabilities. Various research efforts have achieved significant progress in multiple areas, including text, images, and videos \cite{b2,b3,b4}. The affordability and convenience of single-view reconstruction make it a valuable tool across many domains. Advanced algorithms and computational power enable the reconstruction of a 3D model from a single image. This method is especially useful in situations where obtaining multi-angle images is challenging. However, single-view point cloud reconstruction methods frequently depend on costly CAD models \cite{b1} and specific priors \cite{b7}, which limit their generalization and scalability.

To address these limitations, researchers are continuously exploring new methods and architectures to improve the effectiveness and application range of single-view point cloud reconstruction. MCC \cite{b5} developed a simple yet effective large-scale learning framework utilizing vanilla transformers. They directly trained the positions of 3D points, extending the reconstruction targets to various objects and scenes. NU-MCC \cite{b6} enhanced MCC's approach by introducing a lightweight transformer variant to reduce computational costs and employing a flexible hybrid representation to improve geometric and texture details. Notably, both works reproject the RGB-D image back to visible points and use some of these points as query points to encode image features through transformers. This innovative method uses partial points to reflect the whole. However, previous methods failed to capture the relationship between local and overall structures, as they could not represent the hierarchical, tree-like nature of the objects forming the point clouds \cite{b8}.
\begin{figure}
    \centering
    \includegraphics[width=1.0\linewidth]{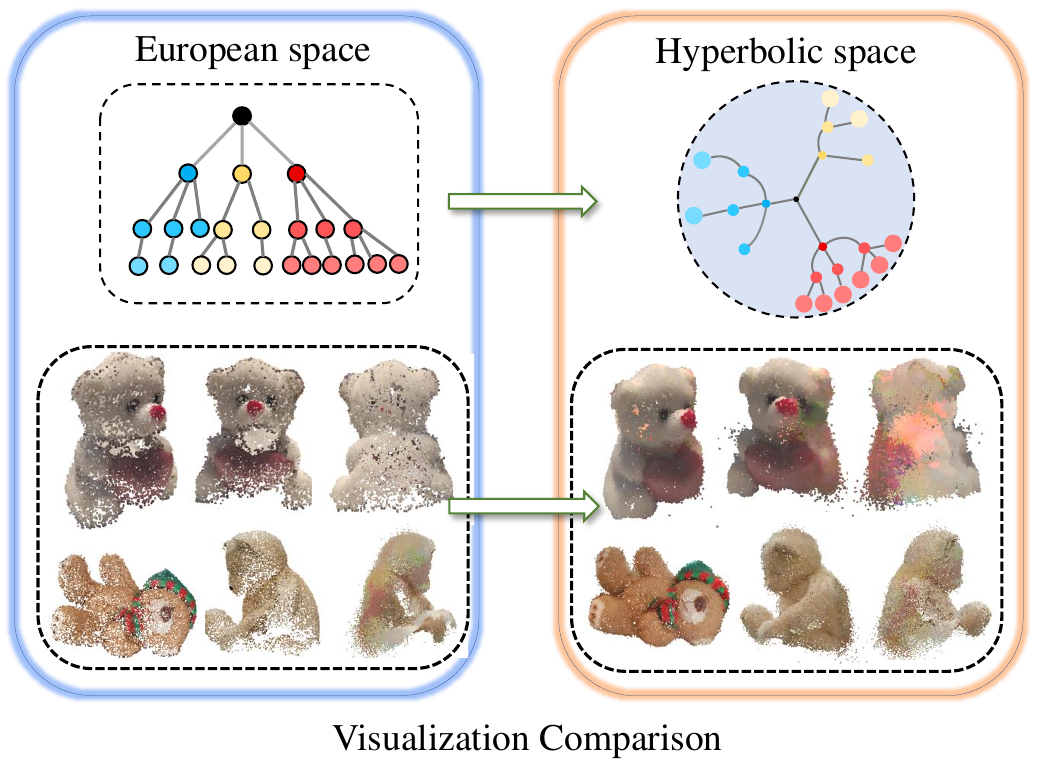}
    \caption{This figure illustrates the efficient embedding of a tree structure from Euclidean space into hyperbolic space. The left side depicts a multi-level tree structure in Euclidean space, while the right side shows the projection of hyperbolic space. Nodes of different colors represent various categories. In hyperbolic space, the node distribution more accurately reflects their hierarchical relationships.} 
    \label{fig1}
\end{figure}

Point cloud data exhibits a rich hierarchical structure, where simple geometric shapes such as disks, squares, and triangles serve as fundamental building blocks for more complex forms. These basic shapes are regarded as "universal ancestors" and can be found across various objects. As illustrated in Fig. \ref{fig1}, this hierarchical structure can be effectively represented as a tree. The lower levels of the tree represent commonly used basic geometric shapes, while the leaves correspond to the specific combinations of these shapes in particular objects. For instance, a simple table comprises legs, a tabletop, and a support structure. The legs connect to and support the tabletop via the support structure, forming a complete table. Essentially, this hierarchy takes the form of a tree, where the legs and support structure serve as the root nodes, with the tabletop as the leaf node. Clearly, understanding this internal hierarchical structure is essential for accurately capturing the 3D object. However, when dealing with data featuring complex structures such as hierarchies or tree-like forms, Euclidean space encounters challenges \cite{b10}, \cite{b11}, \cite{b12}, \cite{b13}. Embedding such data in Euclidean space can result in significant distortions \cite{b10}. In hyperbolic space, distance calculations better preserve the intrinsic structure of the data, more accurately representing the true relationships between data points. This limitation arises because the volume in Euclidean space grows polynomially with its radius, rather than exponentially, thereby restricting its capacity to represent tree-like data with an exponential number of leaves \cite{b9}. 

Hyperbolic space offers a natural solution to this issue. We redefined the hyperbolic chamfer distance in hyperbolic space and introduced a regularization loss and a triplet loss to emphasize the relationship between local point clouds and the overall point cloud. Local features can be embedded in lower-dimensional regions of hyperbolic space, while global features occupy higher-dimensional regions. This approach maximizes mutual information between parts and the whole, thereby enhancing the effectiveness of feature extraction. The main contributions can be summarised as follows:

\begin{itemize}
    \item We introduce hyperbolic space into 3D point cloud reconstruction for the first time, facilitating the representation and understanding of complex hierarchical structures within 3D objects with minimal distortion. This method more accurately captures the dynamic relationship between parts and the whole, offering new perspectives and techniques for 3D reconstruction.
    \item We propose the hyperbolic Chamfer distance and regularized triplet loss, which effectively enhance the relationship between partial and complete point clouds. Additionally, we design adaptive boundary conditions to enhance the model's reconstruction of 3D structures.
    \item Our model demonstrates the superiorities among current mainstream models. Ablation studies highlight the significance of our model and its individual components.
\end{itemize}

\section{Related work}
\subsection{Single-view Point Cloud Reconstruction}
A single image provides limited information and is subject to occlusions and viewpoint constraints. Single-view 3D reconstruction is a complex and challenging research area. Traditionally, this task relies on rich supervision information, such as using CAD models \cite{b1}, point clouds \cite{b15} \cite{b16}, or meshes as training targets to predict 3D shapes from a single image. These methods have shown impressive reconstruction results on simple synthetic datasets like ShapeNet \cite{b14} but are usually limited to specific categories of objects, with limited applicability and generality of the reconstruction. With the advancement of technology, researchers have started exploring weakly supervised and category-agnostic 3D reconstruction methods, aiming to learn 3D representations from a broader and more diverse range of real-world data \cite{b17} \cite{b18}. This includes using RGB-D images or multi-view images to train models, allowing them to capture richer geometric details and visual features. For example, the MCC \cite{b5} project learns compact 3D appearance and geometry from large-scale multi-view data, demonstrating generalization capabilities in natural environments. Additionally, the recent NU-MCC \cite{b6} achieves more efficient 3D reconstruction by introducing the Repulsive Unsigned Distance Field and a neighborhood decoder.
\subsection{Local and Global Point Cloud Processing}
In the field of point cloud data processing and analysis, effectively integrating local and global information is crucial for understanding and utilizing 3D geometric data \cite{b19}. Early methods, such as PointNet \cite{b20}, modeled directly on raw 3D coordinate sets, independently processed each point through shared multi-layer perceptrons (MLP), demonstrating the need for point permutation invariance in point cloud processing. However, PointNet failed to fully leverage local structural information of the point cloud, which has been key to the success of convolutional neural networks (CNNs) in 2D image processing. To address this limitation, PointNet++ \cite{b21} introduced a hierarchical feature learning strategy, enhancing model performance by progressively extracting local features at different scales.

Subsequent research, such as PointCNN \cite{b22}, DGCNN \cite{b23}, and PointConv \cite{b24}, treated point clouds as graph structures and explored the importance of local structures by processing points and their neighborhood relationships within the graph convolution framework. These methods effectively captured local relationships between points through dynamically computed k-nearest neighbor graphs or other graph structures, thereby improving feature representation and model generalization. In terms of global feature learning, methods like PointGLR \cite{b25} have attempted to map local features into a representation space with global properties (e.g., hypersphere), achieving an effective fusion of global and local features.
\subsection{Hyperbolic Geometry}
Hyperbolic space, due to its inherent property of encoding the intrinsic hierarchical structure of data, has become a significant direction for constructing structured representations. Hyperbolic space is a Riemannian manifold with constant negative curvature, making it particularly suitable for handling data with prominent hierarchical structures, such as tree-like structures and complex networks. Inspired by Sarkar's work \cite{b26}, which demonstrated that tree structures could be embedded in hyperbolic space with minimal distortion, several studies have since explored how to achieve effective representation learning in this non-Euclidean manifold. In the field of machine learning, especially in natural language processing and graph data analysis \cite{b27,b28}, embeddings in hyperbolic space have been proven to effectively capture the hierarchical structure and complexity of data. For example, hyperbolic neural networks \cite{b29} and Poincarè embeddings leverage the advantages of hyperbolic geometry to improve model performance and interpretability. Hyperbolic embeddings also show potential in encoding the hierarchical structure of visual data, such as zero-shot classification \cite{b13} and audio-visual question answering \cite{b30}. 

Recently, hyperbolic space has been explored for processing point cloud data \cite{b31} \cite{b9}, particularly in expressing the hierarchical attributes of 3D objects. 3D objects naturally exhibit hierarchical structures, with parts of varying sizes forming the entire object, where smaller parts may be shared among different categories of objects, while larger parts are more category-specific. In point cloud object reconstruction, hyperbolic space is more suitable than traditional Euclidean space for handling objects with pronounced hierarchical or branching structures due to its unique negative curvature properties. When reconstructing models with complex branches, hyperbolic space can naturally express the progressive unfolding and branching characteristics of these structures. Additionally, the geometric structure of hyperbolic space supports exponential spatial expansion, allowing for the efficient embedding and representation of highly hierarchical data structures in lower dimensions, avoiding the data sparsity and computational inefficiency issues that arise in Euclidean space with increasing dimensions. This characteristic enables hyperbolic space to precisely capture and represent the intrinsic hierarchy of objects while maintaining computational manageability and efficiency.

\begin{figure*}
	\centering
	\includegraphics[width=1\linewidth]{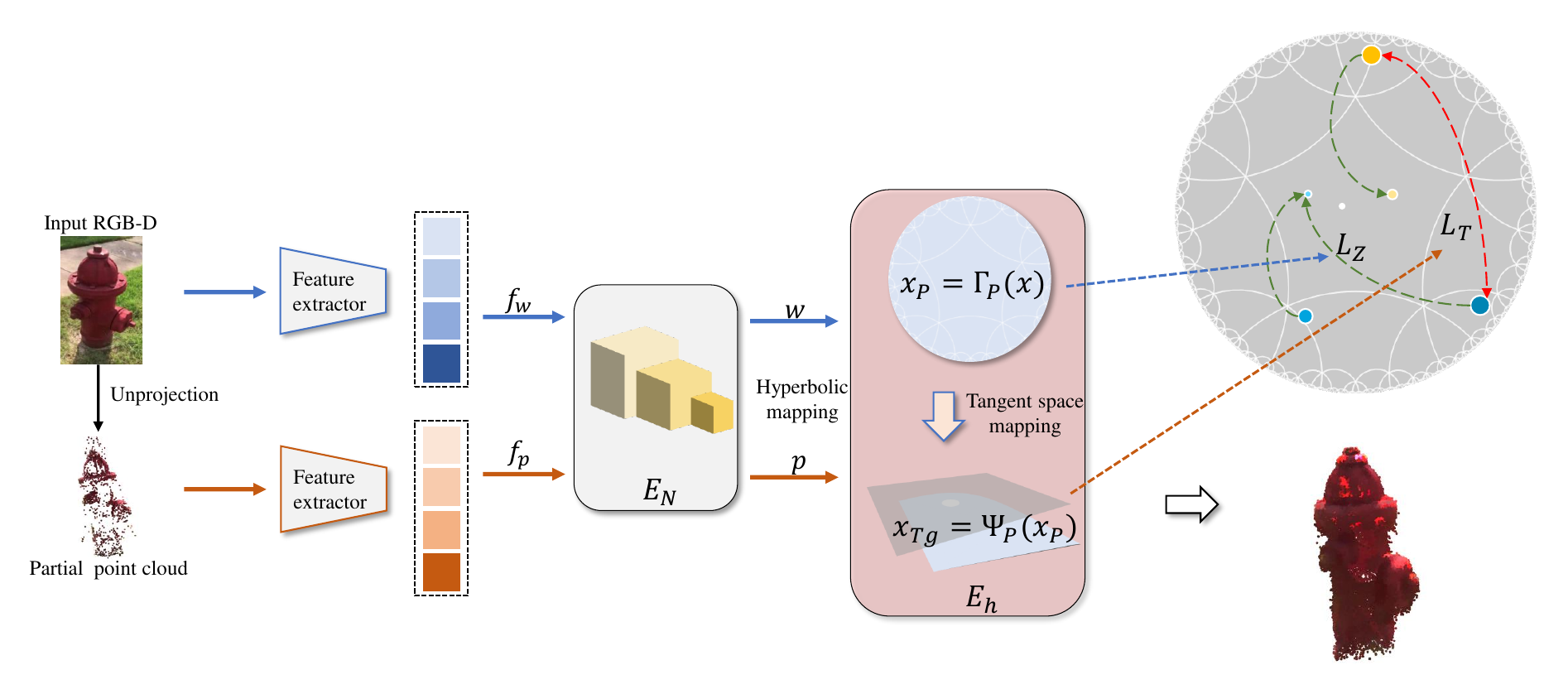}
	\caption{Architecture of the HcPCR. To enhance the understanding of the hierarchical structure of the point clouds, we embed both partial and complete features into a hyperbolic space. The lower-level features (partial point cloud) are embedded into the lower-dimensional central region, while the higher-level features (complete point cloud) are distributed in the higher-dimensional peripheral region. Additionally, a triplet loss is utilized to push apart point clouds of different categories and to pull together point clouds with similar attributes.
}
	\label{fig2}
\end{figure*}
\section{Method}

The overview of HcPCR are illustrated in Fig. \ref{fig2}. Our task is to reconstruct the complete point cloud of an object from a given RGB-D image. For each RGB-D image, we denote the RGB image as \( I \) and the depth map as \( D \). Firstly, we process the image \( I \) into \(16\times16\) patches, and then obtain the global features \( f_w \) using a pre-trained Vision Transformer (ViT) \cite{b32}. For the local features \( f_p \), we perform unprojection of the depth map \( D \) into the 3D world to obtain a patial point cloud of size \( N \times 3 \), where \( N = H \times W \). Due to objective factors like sensor uncertainties, the obtained patial point cloud is redundant and noisy. Therefore, we pool it and encode the local features \( P \) using ViT. Finally, we employ an anchor point prediction operation similar to NU-MCC to further encode the features (\(E_N\)). For specific details about the encoder, please refer to NU-MCC. All losses in NU-MCC are represented as \( \mathcal{L}_N \).
\subsection{Hyperbolic Network}

 Hyperbolic space is a curved space with constant negative curvature ( \(k < 0\)). To transfer feature vectors from Euclidean space to hyperbolic space, we need to perform two stages: hyperbolic mapping and tangent space mapping.

1) \textbf{Hyperbolic Mapping.} In this paper, we use the Poincaré ball model\cite{b33} to represent hyperbolic space. We define an \(n\)-dimensional Poincaré ball with curvature \(k\) as follows:

\begin{equation}
\begin{aligned}
\mathbb{P}_k^n = \{ x \in \mathbb{R}^n : \|x\| < 1/|k| \},\
\end{aligned}
\label{e1}
\end{equation}
where \(\|\cdot\|\) denotes the Frobenius norm.

To project a point \(x \in \mathbb{R}^n\) from Euclidean space \(\mathbb{R}^n\) into an \(n\)-dimensional Poincaré ball \(\mathbb{P}_k^n\) with curvature \(k\), we perform the following hyperbolic mapping:

\begin{equation}
\begin{aligned}
\
x_P = \Gamma_\mathbb{P}(x) = 
\begin{cases} 
x, & \text{if } \|x\| \leq 1/|k| \\
\frac{1 - \varepsilon }{|k|} \frac{x}{\|x\|}, & \text{else}
\end{cases}
\
\end{aligned}
\label{e2}
\end{equation}
where \(x_P\) is the projected point in the Poincaré sphere and \(\varepsilon\) is a scaling factor adjusted according to the curvature \(k\).

2) \textbf{Tangent Space Mapping.} Hyperbolic spaces differ from vector spaces in that they do not themselves support operations such as unusual addition. In order to facilitate such operations in hyperbolic spaces, we must employ Möbius addition \cite{b27}:
\begin{equation}
\begin{split}
z_P \oplus_k x_P &= \frac{(1 + 2|k| \langle z_P, x_P \rangle + |k| \|x_P\|^2) z_P}
{1 + 2|k| \langle z_P, x_P \rangle + |k|^2 \|z_P\|^2 \|x_P\|^2} \\
&\quad + \frac{(1 - |k| \|z_P\|^2) x_P}
{1 + 2|k| \langle z_P, x_P \rangle + |k|^2 \|z_P\|^2 \|x_P\|^2},
\end{split}
\label{e3}
\end{equation}
the hyperbolic tangent space is a local Euclidean space that approximates the hyperbolic space. To facilitate the calculation of loss, we project points from hyperbolic space into the hyperbolic tangent space. Given a point \( z_P \in \mathbb{P}_k^n \) in the Poincaré sphere, we can generate a tangent plane \(T_z\mathbb{P}_k^n\):

\begin{equation}
\begin{aligned}
x_{Tg} = \Psi_\mathbb{P}(x_P) = \frac{2}{\sqrt{|k| \lambda_k(z_P)}} &\tanh^{-1}\left(\sqrt{|k|} \|  - z_P \oplus_k x_P \| \right) \\
& \cdot \frac{-z_P \oplus_k x_P}{\| -z_P \oplus_k x_P \|},
\end{aligned}
\label{e4}
\end{equation}
where \(\langle , \rangle\) denotes the inner product and \(x_{Tg}\in T_z\mathbb{P}_k^n\). To facilitate understanding and analysis, we set \( z_P = 0_P \) .
\subsection{Hyperbolic distance}
Point cloud distance is a method used to measure the dissimilarity between two point cloud sets. In practical applications, due to the unordered nature of point clouds, shape-level distance typically relies on statistical analysis of point-to-point distances. Among these, Chamfer Distance (CD) is widely used due to its relatively efficient computation  \cite{b35} \cite{b34}. CD evaluates the dissimilarity between two point clouds by calculating the average distance from each point in one point cloud to the nearest point in the other point cloud. Generally, the Chamfer distance of point clouds is defined as follows:

\begin{equation}
\begin{aligned}
 D(x_i, y_i) = \frac{1}{|x_i|} \sum_j \min_k d(x_{ij}, y_{ik}) + \frac{1}{|y_i|} \sum_k \min_j d(x_{ij}, y_{ik}) .
\end{aligned}
\label{e5}
\end{equation}

In Euclidean space, we typically define the function \(d\) as:

\begin{equation}
d(x_{ij}, y_{ik}) = 
\begin{cases} 
\|x_{ij} - y_{ik}\| & \text{as L1-CD}; \\
\|x_{ij} - y_{ik}\|^2 & \text{as L2-CD},
\end{cases}
\label{e6}
\end{equation}
where \(\| \cdot \|\) denotes the Euclidean \( \ell_2 \) norm of the vector.

Although Chamfer Distance (CD) performs well in computing point cloud distances, the gradient of the CD distance assigns higher weights to points with larger distances, making it sensitive to outliers\cite{b31}. Hyperbolic space offers greater flexibility than Euclidean space in measuring distances between points, potentially correcting matching errors in CD. Therefore, we introduce a new distance metric, the Hyperbolic Chamfer Distance (HyperCD):

\begin{equation}
\begin{aligned}
d(\mathbf{x}, \mathbf{y}) = \frac{2}{\sqrt{k}} \text{arctanh} \left(\sqrt{k}\|-\mathbf{x} \oplus_k \mathbf{y}\|\right).
\end{aligned}
\label{e7}
\end{equation}

It is worth mentioning that when \( k \rightarrow 0 \), the hyperbolic distance can be simply expressed as the Euclidean distance: 

\begin{equation}
\begin{aligned}
\lim_{k \rightarrow 0} d(\mathbf{x}, \mathbf{y}) = 2\|\mathbf{x} - \mathbf{y}\|.
\end{aligned}
\label{e8}
\end{equation}

\subsection{Hyperbolic Point Cloud loss}

During the training of HcPCR, we defined a loss function \(\mathcal{L}\) that consists of three different components: the NU-MCC loss \(\mathcal{L}_N\), the regularization loss \(\mathcal{L}_Z\), and the triplet loss \(\mathcal{L}_T\):
\begin{equation}
\begin{aligned}
\mathcal{L} & = \mathcal{L}_N + \mathcal{L}_Z + \mathcal{L}_T .
\end{aligned}
\label{e9}
\end{equation}
\subsubsection{Regularisation loss.}
As introduced in the previous section, we need to project \( w \) and \( p \) from Euclidean space to hyperbolic space. We use a hyperbolic mapping (Eq. \eqref{e2}) to map Euclidean feature vectors into hyperbolic space, and then obtain the hyperbolic embeddings of the entire point cloud \( W \) through tangent space mapping (Eq. \eqref{e4}). For the partial point cloud \( P \), we repeat the same process.

To achieve part-whole hierarchy, we define the following regularization loss:

\begin{equation}
\begin{aligned}
\mathcal{L}_{Z} = \max \left( 0, -\Gamma_\mathbb{P}(\mathcal{W} ^+) + \Gamma_\mathbb{P}(\mathcal{P} ^+) + \frac{\gamma}{N} \right), 
\end{aligned}
\label{e10}
\end{equation}
where \(\mathcal{W}^+\) and \( \mathcal{P}^+\) are the hyperbolic representations of the whole and part of the same point cloud.
\begin{table*}[t]
	\centering
	\begin{threeparttable}
		\caption{The performance comparison on CO3D-v2 datasets.}
		\label{TAB1}
		\fontsize{9}{14}\selectfont 
		\setlength{\tabcolsep}{12pt} 
		\begin{tabular}{c|c|ccccc|c}  
			\toprule[1.5pt]  
			Method & Distance & Acc↓ & Com↓ & Pre↑ & Recall↑ & \(L_1\)↓ & F1↑ \\ 
			\hline
 			MCC & Euclidean  & 0.172 & 0.144 & 0.689 & 0.727 & 0.316  &0.739 \\
 			NU-MCC(base) & Euclidean  & 0.121 & 0.146 & 0.792 & 0.840 & 0.267  & 0.802 \\
 \hline
			baseline+\(L_Z\) & Euclidean & 0.149 & 0.152 & 0.763 & 0.813 & 0.301  & 0.786  \\ 
   			baseline+\(L_T\)& Euclidean & 0.119 & 0.134 & 0.817 & 0.835 & 0.253  & 0.814  \\ 
   			baseline+\(L_Z\)+\(L_T\)& Euclidean & 0.127 & 0.139 & 0.728 & 0.853 & 0.266  & 0.806  \\ 
       \hline
        \rowcolor{gray!20}
 			baseline+\(L_Z\) & Hyperbolic  & 0.112 & 0.108 & 0.827 & 0.853 & 0.220  & 0.836 \\
     \rowcolor{gray!20}
 			baseline+\(L_T\) & Hyperbolic  & 0.107 & 0.101 & 0.832 & 0.868 & 0.208  & 0.853 \\
     \rowcolor{gray!20}
  			baseline+\(L_Z\)+\(L_T\)(ours) & Hyperbolic  & \textbf{0.104} & \textbf{0.094} & \textbf{0.848} & \textbf{0.884} & \textbf{0.198}  & \textbf{0.861} \\
			\bottomrule[1.5pt]
		\end{tabular}
	\end{threeparttable}
\end{table*}

The regularizer \(\mathcal{L}_{Z}\) in Eq. \eqref{e10} introduces a part-whole hierarchy by encouraging partial embeddings to be closer to the center of the Poincaré ball, while the overall embeddings are closer to the edge. Inspired by \cite{b9}, we employed a variable margin \( \gamma/N \) that depends on the number of points \( N \) in a part \( P \). This means that simpler shapes, composed of fewer points, will be farther from the representation of the whole object and have a lower hyperbolic norm (i.e., closer to the center). In contrast, embeddings of larger parts will gradually approach the edge of the Poincaré ball based on their size. 

Relying on a fixed margin defined by hyperparameters might not effectively handle the complex structure of point cloud data, thereby affecting the quality of overall and partial feature representations. To address this issue, we introduced adaptive margins by incorporating whole-part features into the hyperbolic manifold. Specifically, we first concatenate the features \( p \) and \( w \) into a vector \( l  \), and then replace the margin with a linear layer (MLP). The detailed method is as follows:

\begin{equation}
\begin{aligned}
\ \gamma = \gamma_0 \cdot \text{sigmoid}(\text{MLP}(l)) .\
\end{aligned}
\label{e11}
\end{equation}
where \( \text{MLP}(\cdot) \) denotes a fully connected layer, and \( \gamma_0 \) represents the initial margin.

\subsubsection{Triplet loss.}
To ensure that parts from different classes maintain a larger geodesic distance, we aim to pull parts and wholes from the same class closer together while mapping parts from different classes further apart. We define the following triplet loss:
\begin{multline}
\mathcal{L}_T = \max(0, d(\Psi_\mathbb{P}(\mathcal{W} ^+), \Psi_\mathbb{P}(\mathcal{P} ^+)) \\
- d(\Psi_\mathbb{P}(\mathcal{W} ^+), \Psi_\mathbb{P}(\mathcal{P} ^-)) +\varepsilon )
\label{e12}
\end{multline}
where \( \varepsilon  \) is a hyperparameter used to control the degree of separation between positive and negative samples , while \(\mathcal{P} ^-\) is the embedding of a part from a different category. 

The triplet loss \(\mathcal{L}_T\) in Eq. \eqref{e12}. encourages the correct clustering of objects and parts in hyperbolic space. Specifically, parts and wholes from the same class are brought closer together, while parts from different classes are mapped farther apart. This ensures that parts from different classes maintain greater geodesic distances. \(\delta\) is a hyperparameter that controls the separation degree between positive and negative samples.

\section{Experiments}
\subsection{Dataset}
We conducted object reconstruction experiments primarily on the CO3D-v2 dataset \cite{b17}. To evaluate the performance of our HcPCR model, we compared it with the latest baseline models considered the best in this field. For consistency, we followed the dataset setup as described in \cite{b5}. 

CO3D-v2 is a dataset used in computer vision research for 3D object recognition and reconstruction. It is an extended version of the original CO3D dataset, featuring more videos and object categories. The dataset consists of videos captured mainly in natural environments, providing multi-angle views. It includes approximately 37,000 short videos covering 51 object categories. The videos include object segmentation masks, and 3D point clouds and depth maps automatically extracted using the COLMAP software. We followed the NU-MCC setup, using 10 categories for evaluation and the remaining 41 for training.

\subsection{Evaluation metrics}
To fair comparison, we adopted the evaluation method proposed by \cite{b5,b6,b36}. The evaluation metrics include the L1 distances for accuracy (Acc) and completeness (Comp), their sum as the Chamfer distance (CD), precision (Prec), recall, and the F-score, which is the harmonic mean of precision and recall. For detailed formulas refer to \cite{b36}.

\subsection{Implementation details}
We adopted the feature extraction method described in NU-MCC \cite{b6}. The input image resolution is 224×224. We used ViT to encode \( P \) and \( I \), generating 768-dimensional feature vectors. Additionally, we followed the data augmentation strategy described in [MCC], applying random scaling and random rotation along each axis. The HcPCR model was trained on an Nvidia A100 GPU. For HcPCR, we set the curvature \( k \) to -0.14, \( \alpha \) to 2.0, \( \gamma_0 \) to 1000, and \( \varepsilon  \) to 4. The initial learning rate was set to 0.0001. We used the Adam optimizer and trained for 100 epochs.
\begin{figure*}[ht]
	\centering	\includegraphics[width=0.9\linewidth]{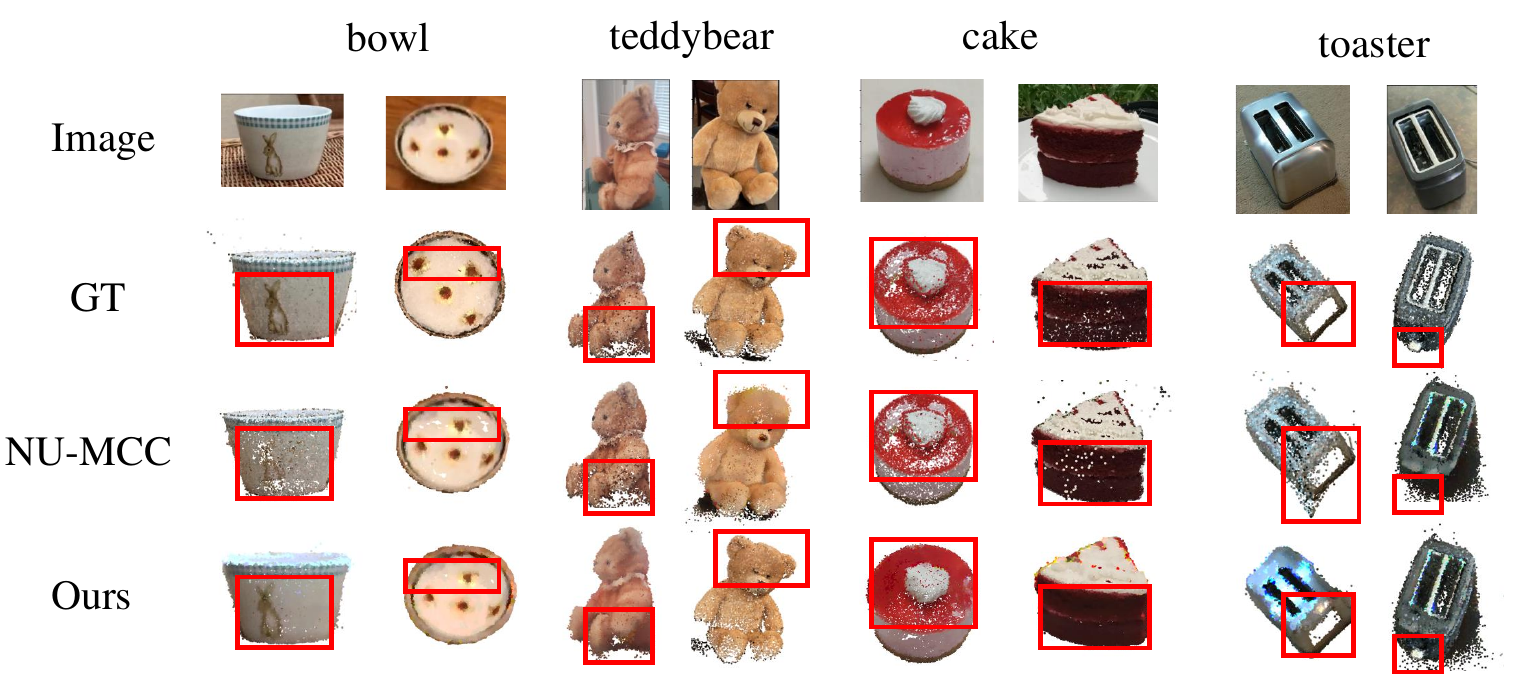}
	\caption{Visualisation comparison on CO3D-v2 validation set.  We have selected three categories. Each category contained two samples for evaluation.}
	\label{fig3}
\end{figure*}

\subsection{Results Comparison}
We compared the HcPCR model with baseline models, including MCC and NU-MCC, on the CO3D-V2 dataset. Table \ref{TAB1} demonstrates that the HcPCR model outperforms all baseline models across every aspect. Specifically, the HcPCR model shows superior performance in all domains compared to NU-MCC. Our method achieves average improvements of 5.9\%, 5.2\%, and 5.6\% in F1 score, completeness, and accuracy, respectively. These results indicate that the HcPCR has superior capability in point cloud reconstruction, attributed to its ability to hierarchically explore 3D structures by leveraging the properties of hyperbolic space.

To investigate the necessity of regularization loss and triplet loss in hyperbolic space, we trained the NU-MCC model using different loss functions. When using Euclidean distance, adding triplet loss effectively enhances the model's expressiveness. However, merely adding regularization loss leads to an overall decline in model performance. This occurs because the regularization term was originally designed to enhance repulsion between complete and partial point clouds, aiming for a hierarchical embedding of the point cloud structure. However, since Euclidean space cannot effectively represent the tree-like structure of point clouds, this operation instead distances or even severs the original connection between partial and complete point clouds. This further demonstrates the inherent advantage of hyperbolic space in handling tree-structured data. Conversely, when using hyperbolic distance as the metric, the addition of both regularization and triplet losses significantly improves model performance. As shown in Table \ref{TAB1}, the model achieves its best performance when using the full loss function in hyperbolic space.
\subsection{Ablation Study}

\subsubsection{Evaluating the effectiveness of different margins \(\gamma\).}
\begin{figure}
	\centering	
    \includegraphics[width=0.45\linewidth]{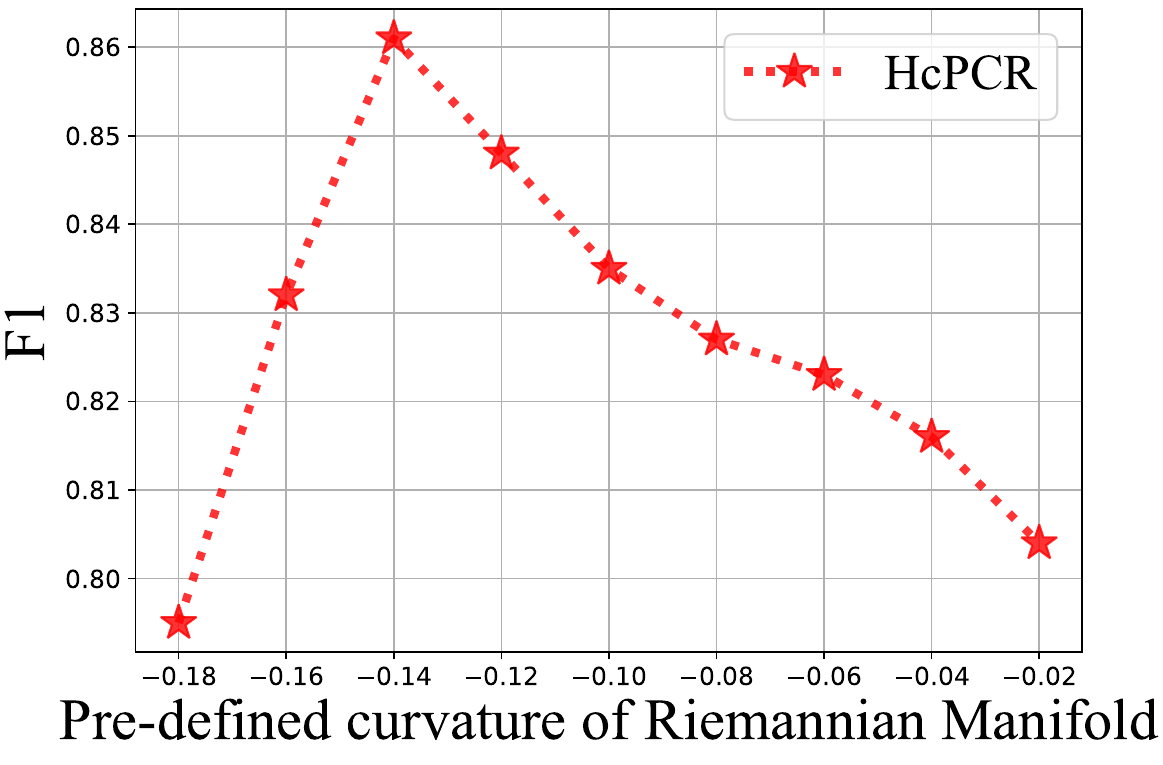}
    \includegraphics[width=0.45\linewidth]{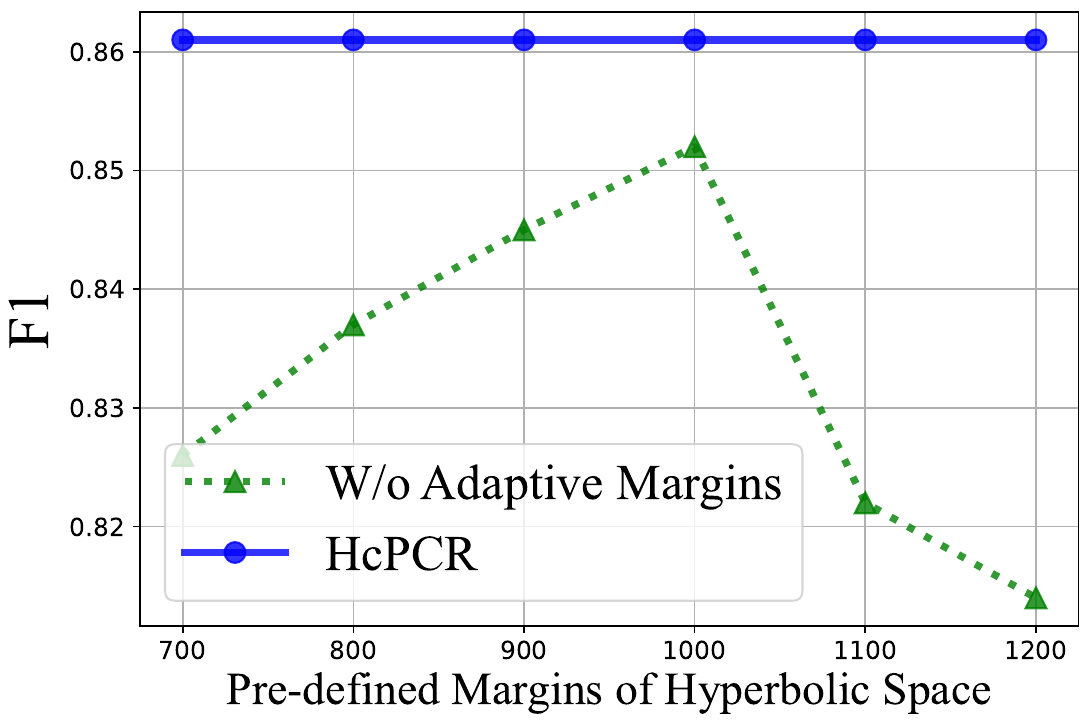}
	\caption{Ablation Study of different margin and curvature.}
	\label{fig6}
\end{figure}

To validate the superiority of adaptive edges, we compared different preset margin values with the adaptive margins component. The experimental results are shown in left of Fig. \ref{fig6}. It is evident that using different preset values without the adaptive margins significantly affects model performance. This observation highlights the necessity of the adaptive margins component.

\renewcommand{\dblfloatpagefraction}{.9}
\begin{figure*}[ht]
	\centering	\includegraphics[width=0.9\linewidth]{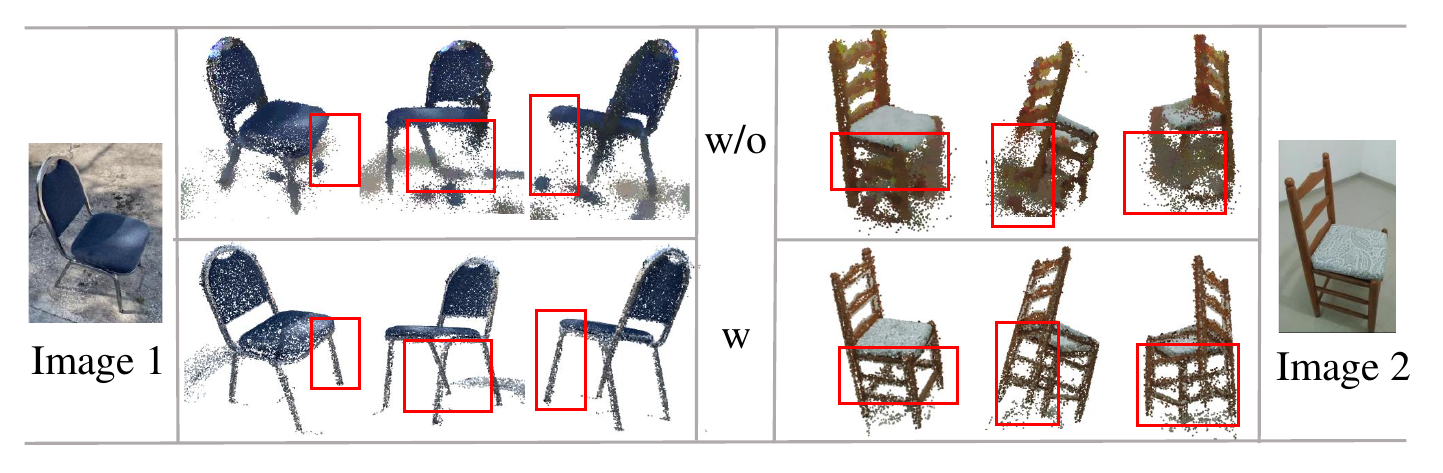}
	\caption{Visualisation results of ablation study with (w) or without (w/o) the full loss function.}
	\label{fig4}
\end{figure*}
\begin{figure}[ht]
	\centering	\includegraphics[width=0.9\linewidth]{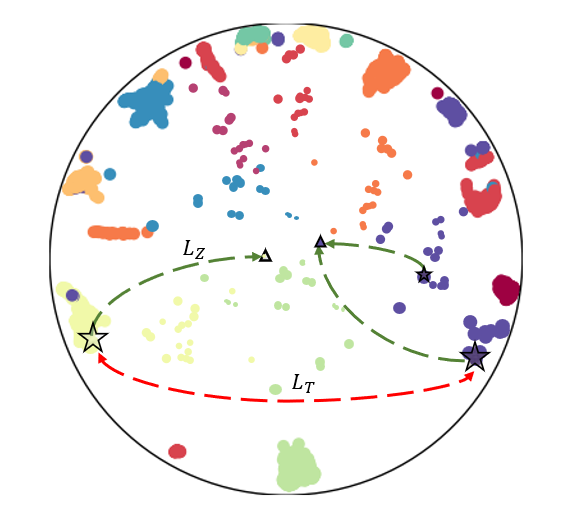}
	\caption{The HcPCR embedding is used to represent the CO3D-v2 dataset on the Poincaré disk. Each point within the disk corresponds to a sample, with different colors representing different categories.}
	\label{fig5}
\end{figure}

\subsubsection{Evaluate the effectiveness of the model under different curvatures \(k\).}
We conducted experiments under different curvatures to understand the impact of hyperbolic geometry on the performance of the model with hyperbolic embeddings. The model's performance was tested under various curvatures, and the results are shown in the right of Fig. \ref{fig6}. The best performance was achieved at a curvature of -0.14. This indicates that this value provides the optimal hyperbolic embedding effect.

\subsubsection{Ablation study for \(\delta\)-hyperbolicity.}
While processing point cloud datasets, we observed that their intrinsic geometric properties often display significant non-Euclidean characteristics due to their inherently complex hierarchical structures. In Table \ref{tab2}, we evaluated the point cloud embeddings extracted by a basic encoder and a fully enhanced hyperbolic network encoder to assess their effectiveness in revealing these hierarchical features. To this end, we employed the \(\delta\)-hyperbolicity measure to evaluate the extent of hierarchical structures, a method previously shown to effectively detect hierarchical characteristics \cite{b27}. The \(\delta\) values computed in the experiments evaluated the overall similarity of sample features to an ideal tree-like structure, where \(\delta\) values close to 0 indicate stronger hyperbolicity. 
\begin{table}
\centering
\setlength{\tabcolsep}{22pt}
\begin{tabular}{lcccc}
\hline
          Encoder &\(\delta\)\  \\ \hline
\(E_{N}\)     & 0.326      \\
\rowcolor{gray!20}\(E_{N}+E_h\)    & 0.294      \\ \hline
\end{tabular}
\caption{The \(\delta\)-hyperbolicity value measure the non-Euclidean geometric properties of the data. A lower \(\delta\) value indicates the data exhibits more pronounced hyperbolicity.}
\label{tab2}
\end{table}
\subsubsection{Visualization on the CO3D-v2 dataset.}
Fig. \ref{fig3} shows the visual comparison of the reconstructed structures by different models. This figure demonstrates that HcPCR significantly enhances the reconstruction of the overall contours. Compared to the baseline, the geometric appearance of the reconstructed objects is much clearer. This improvement could be attributed to the hyperbolic constraint, which preserves the details of certain point clouds from the original data. Additionally, our reconstructed 3D structures are noticeably more complete.

\subsubsection{Visualization for ablation study on loss function. }
To further validate the effectiveness of the hyperbolic geometry constraint, we present visual results using the full loss function and using only the initial \( L_Z \) loss, in Fig. \ref{fig4}. It is evident that, with the addition of the hyperbolic geometry constraint, the model generates more refined and detailed textures. Notably, for geometrically prominent parts of the object, the reconstructed shapes are more streamlined and complete. Additionally, our model demonstrates stronger generalization ability. Under the hyperbolic geometry constraint, the reconstruction of occluded object structures (highlighted in the red box) in the images still performs remarkably well.

\subsubsection{The hierarchical structure of point cloud data in hyperbolic space.}
We used UMAP \cite{b37} to project high-dimensional point clouds onto a two-dimensional Poincaré disk, illustrating how hyperbolic space explores the hierarchical structure of point cloud data, as shown in Fig. \ref{fig5}. We visualized 10 categories, each containing 3-4 partial point clouds and one complete point cloud. The number of points in the partial clouds gradually increases, causing these partial objects to move closer to the edge of the disk. As more points are added and the partial objects gradually combine into a complete object, their hyperbolic norm increases, ultimately pushing the entire object towards the disk's edge. This process highlights the impact of hyperbolic space on hierarchical embedding.

\section{Conclusion}
We propose a hyperbolic neural network for reconstructing point cloud objects. To enhance the hierarchical structure of point clouds, we introduce regularization and a triplet loss in hyperbolic space. Additionally, we design an adaptive margins component within the regularization loss to mitigate the heterogeneity caused by fixed margin values. Furthermore, we conduct an ablation study to verify the effectiveness of our model. Finally, we visualize how hierarchical embeddings of tree-like data are achieved in hyperbolic space and compare the reconstruction results against other models.
\section{Acknowledgments}
This work was supported in part by the National Key R\&D Program of China (2021YFF0900500), the National Natural Science Foundation of China (NSFC) under grants 62441202, U22B2035, 20240222, and the Fundamental Research Funds for the Central Universities under grants HIT.DZJJ.2024025.

\end{document}